\pdfoutput=1
\documentclass[11pt]{article}

\usepackage[]{acl}

\usepackage{times}
\usepackage{latexsym}

\usepackage[T1]{fontenc}
\usepackage[utf8]{inputenc}

\usepackage{microtype}

\usepackage{inconsolata}

\usepackage{graphicx}
\usepackage{booktabs}
\usepackage[symbol]{footmisc}
\usepackage{authblk}
\let\ul\underline
\newcommand\blfootnote[1]{
    \begingroup
    \renewcommand\thefootnote{}\footnote{#1}
    \addtocounter{footnote}{-1}
    \endgroup
}

\title{CADReN: Contextual Anchor-Driven Relational Network for Controllable Cross-Graphs Node Importance Estimation}

\author[1 $\dagger$]{Zijie Zhong }
\author[2 $\dagger$]{Yunhui Zhang }
\author[2]{Ziyi Chang }
\author[1 *]{Zengchang Qin }
\affil[1]{ Beihang University}
\affil[2]{ theSight Technology}

\begin{document}
\maketitle

\blfootnote{$\dagger$ Both authors contributed equally}
\blfootnote{$*$ Corresponding author: zcqin@buaa.edu.cn}

\begin{abstract}
Node Importance Estimation (NIE) is crucial for integrating external information into Large Language Models through Retriever-Augmented Generation. Traditional methods, focusing on static, single-graph characteristics, lack adaptability to new graphs and user-specific requirements. \textbf{CADReN}, our proposed method, addresses these limitations by introducing a Contextual Anchor (CA) mechanism. This approach enables the network to assess node importance relative to the CA, considering both structural and semantic features within Knowledge Graphs (KGs). Extensive experiments show that CADReN achieves better performance in cross-graph NIE task, with zero-shot prediction ability. CADReN is also proven to match the performance of previous models on single-graph NIE task. Additionally, we introduce and opensource two new datasets, \textbf{RIC200} and \textbf{WK1K}, specifically designed for cross-graph NIE research, providing a valuable resource for future developments in this domain.

% \textbf{Keywords}: Node Importance Estimation, Node Classification, Knowledge Graph, Cross-attention
\end{abstract}

\section{Introduction}

% With the rapid development of Transformer\citep{transformer}-based Large Language Models (LLMs) \citep{gpt2, gpt3, gpt4, llama}, more and more effort have been put into building AI Agents that could conduct high-quality analysis, make independent decisions and generate insights in real-world production scenario. However, relying solely on LLMs is not sufficient to achieve a satisfactory due to LLMs' problem of "hallucination". Aiming to solve this, Retriever-Augmented Generation (RAG) \citep{rag} rose to be an important paradigm of design of the modern AI Agents. The main idea of RAG is to integrate external information to the LLM, in order to enhance its performance. In the direction of RAG, Knowledge Graphs (KGs) are put under the spotlight. Unlike the information stored as plain text or images, the knowledge represented in KGs are well structured and precise.  

The advent of Transformer-based Large Language Models (LLMs) \citep{transformer, gpt2, gpt3, gpt4, llama} has catalyzed the development of AI Agents for advanced analytical and decision-making tasks. Yet, LLMs alone are prone to "hallucination," leading to inaccuracies. The introduction of Retriever-Augmented Generation (RAG) \citep{rag} has become essential to enhance LLMs by integrating structured and precise Knowledge Graphs (KGs), thereby mitigating this issue.

\begin{figure}[t]
  \centering
  \resizebox{\columnwidth}{!}{
  \includegraphics{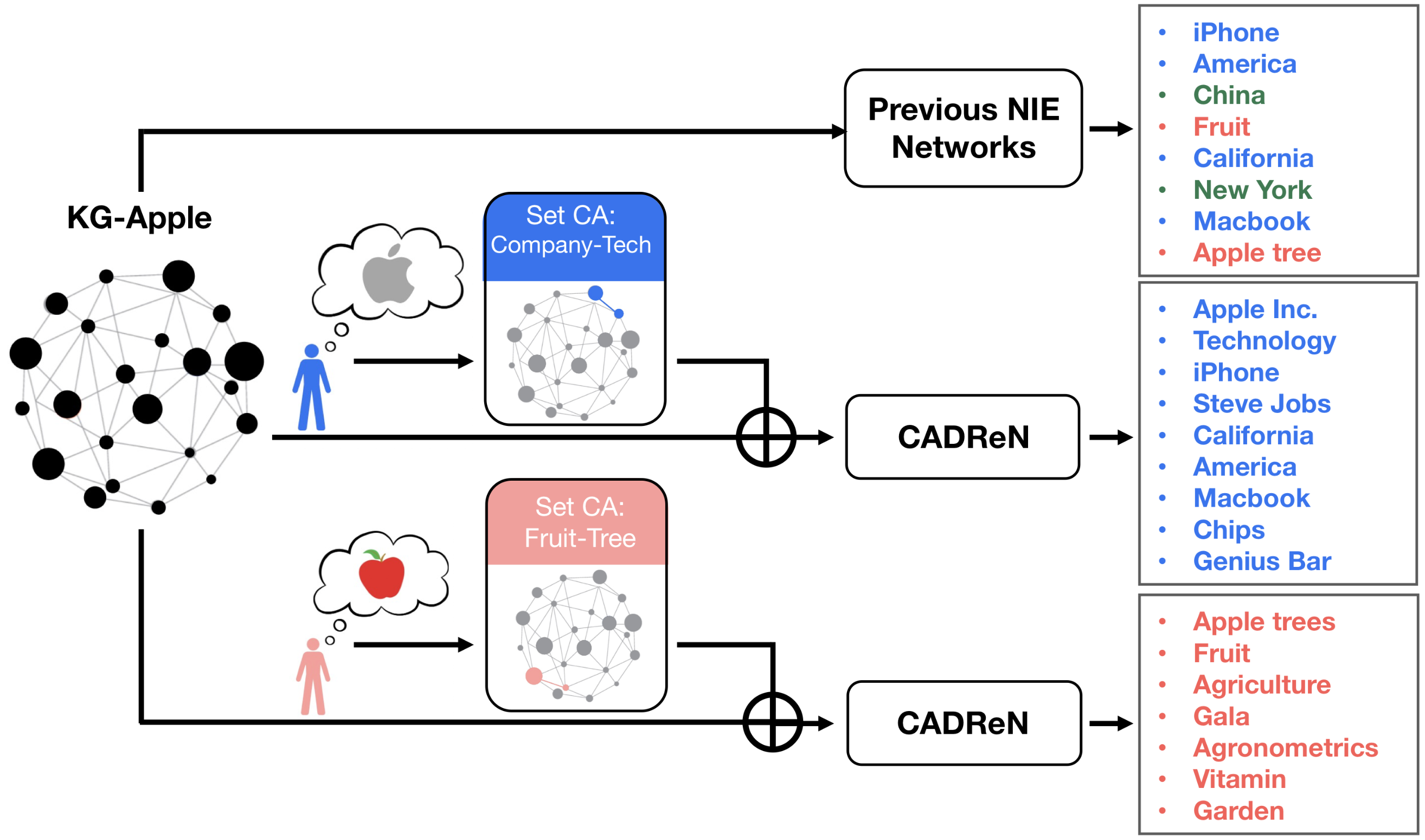}}
  \caption{
  CADReN leverages user-defined Contextual Anchors (CAs) to enhance precision in KG queries. In the figure, \textit{KG-Apple} contains diverse information related to \textit{Apple}. Users applying \textit{Company-Tech} and \textit{Fruit-Tree} CAs receive focused outputs via CADReN, contrasting with the generalized results given by previous NIE networks without CA utilization."
  }
  \label{fig:showcase}
\end{figure}

\begin{figure*}[t]
  \centering
  \resizebox{\textwidth}{!}{
  \includegraphics{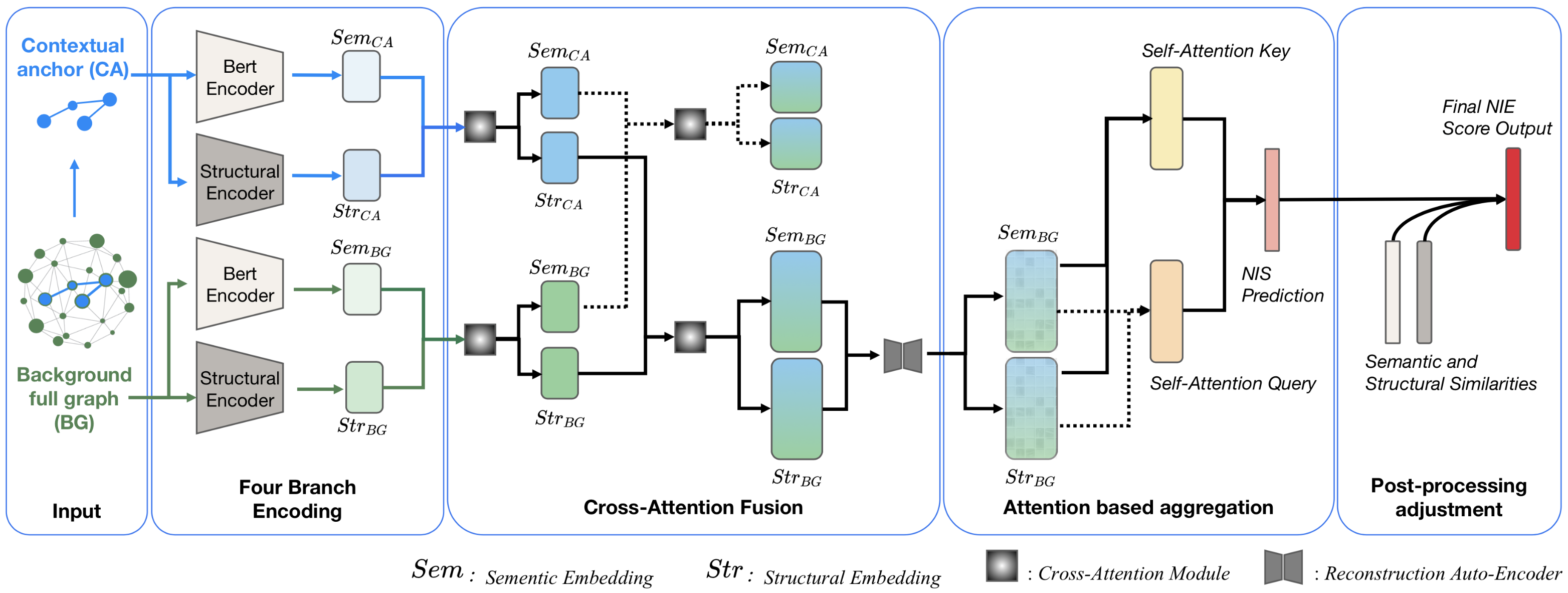}}
  \caption{The figure above presents the overall architecture of the CADReN model. The semantic and structural information in CA and BG are encoded in BERT and our proposed structural encoder, respectively. Cross-attention fusion is then applied to the combinations of these embeddings to capture the relational information between CA and BG. The BG embeddings mixed with the information from CA are then used to predict the NIE scores, with the introduction of Reconstruction Auto-encoder, Attention-based Aggregation mechanism and Post-Processing mechanism to improve the quality of the output.}
  \label{fig:pipeline}
\end{figure*}

% KGs provide a structured representation of knowledge and serve as an expandable data format to store, retrieve and update heterogeneous data. Through KGs, complex relationships among entities are being mapped. The utilization of structured knowledge facilitates the recognition of patterns and the formation of insights. Thanks to the high performance graph management techniques like Neo4j \citep{neo4j}, KGs have been widely adopted in fields that are highly relied on structural information, such as recommendation system \citep{personalizereco},  fraud risk management \citep{infdetect}, and medicine discovery \citep{geometric, geometricmolecular}. KGs have been proven to be a useful tool to inject external knowledge to LLMs to enhance them.

KGs provide a structural framework to encapsulate heterogeneous data, allowing for intricate mappings of entity relationships. Their structured nature is conducive to pattern recognition and insight formation. Enhanced by high-performance graph management systems such as Neo4j \citep{neo4j}, KGs have become integral to domains dependent on structural information, including recommendation systems \citep{personalizereco}, fraud detection \citep{infdetect}, and drug discovery \citep{geometric, geometricmolecular}. Their structured knowledge is essential for augmenting LLMs to improve performance.

% In order to enhance AI's usage within the business sphere by empowering AI agents with the capabilities to conduct tasks like identifying new business opportunities  and predicting possible disruptive events/factors, an increasing number of researches have been conducted by combining LLMs with KGs \citep{unifyingllmkg}. Through these projects we learn that the performance of KG-augmented LLM is largely determined by the quality of information retrieval. The process of retrieving relevant and pertinent information from a KG has been formulated as Node Importance Estimation (NIE) task, and has gained more and more attention.

Within the business sphere, leveraging AI to identify new opportunities and predict market disruptions has become a research focus. Integrating KGs with LLMs \citep{unifyingllmkg} has proven critical, with the effectiveness of KG-enhanced LLMs heavily reliant on the quality of retrieved information. This retrieval, defined as the Node Importance Estimation (NIE) task, is increasingly recognized for its significance.

% NIE is a vital task in Information Retrieval, aiming to assign an importance score to each node within a graph. With this importance score, the most relavant information in a KG could be easily distinguished, facilitating the following Information Retrieval and RAG. To tackle NIE, recent works primarily focused on two approaches: Structure-Pattern-Based Methods, such as PageRank \citep{Page1999ThePC},  HITS \citep{hits} and HAR \citep{har}, and Embedding Based Methods, such as  GNN \citep{gnn, understandinggnn}, GENI \citep{geni} and RGTN \citep{RGTN}. Nevertheless, these methods have some critical defaults that hinder their applications on real-world scenarios. First, they are designed specifically for learning the absolute information inside single static graph, unable to migrate the knowledge learned from one graph to another. In other words, they need to be retrained on each different graph. Second, previous works have one fixed criterion of "importance" once they are trained. The model evaluates the importance score of each node in a black-box, uncontrollable way, which often results in a inconsistency with the users' interest (as illustrated in Fig.~\ref{fig:showcase}).

NIE is a fundamental aspect of Information Retrieval, focusing on evaluating and scoring the relevance of nodes in a Knowledge Graph. This process plays a crucial role in enhancing the effectiveness of RAG by ensuring the most pertinent graph information is prominently featured. Current approaches, including Structure-Pattern-Based Methods like PageRank \citep{Page1999ThePC}, HITS \citep{hits}, HAR \citep{har}, and Embedding-Based Methods like GNN \citep{gnn, understandinggnn}, GENI \citep{geni}, and RGTN \citep{RGTN}, are hindered by two major deficiencies: their focus on static single-graph information and the inability to transfer learning across graphs without retraining. Additionally, their static definition of "importance" often leads to outputs that may not align with the specific interests of users. (see Fig.~\ref{fig:showcase}).

% To solve these drawbacks, we proposed a novel approach, CADReN (\textbf{C}ontext \textbf{A}nchor \textbf{D}riven \textbf{Re}lational \textbf{N}etwork) for cross-graphs NIE tasks. Contextual Anchor (CA) are a set of nodes mentioned in the user's input query, which indicates users' key focus and interests. CADReN is trained to model the relational relations between the CA nodes and the rest of the KG, and the node importance scores are computed based on the CA nodes. The whole architecture could be divided into 4 parts: Four Branch Encoding, Cross-Attention Fusion, Attention-based Aggregation, and Post-processing Adjustment, as explained in Fig.~\ref{fig:pipeline}. Intrinsically, our network is designed to capture the relative relational information between the nodes, making it transferable from one graph to another. The CA nodes are introduced as an ``anchor" when assessing the importance scores, who always have the highest importance scores and the importance scores of other nodes are calculated based on their "related importance" compared to the CA nodes. CA nodes, or the users' input query from which they are extracted, function also as an interface between users and the whole network, allowing users to adjust the outcome and retrieve the information that is better aligned to their needs (as illustrated in Fig.~\ref{fig:showcase}). Extensive experiments showed the effectiveness of our method, especially on multi-graph tests. The detailed architecture of our network and the experiment results will be discussed in detail in subsequent chapters.

Addressing these challenges, we introduce \textbf{CADReN} (\textbf{C}ontext \textbf{A}nchor-\textbf{D}riven \textbf{Re}lational \textbf{N}etwork) for cross-graph NIE tasks. CADReN leverages user input—Contextual Anchors (CA)—to delineate relative node importance within the KG, enabling transferability across graphs and user-driven result customization (detailed in Fig.~\ref{fig:pipeline}).  Extensive experiments showed the effectiveness of our method, especially on multi-graph tests.

% The subsequent sections of this paper are organized as follows: firstly, we present a comprehensive review of related works in the field of NIE; subsequently, we give a formal definition of the core concepts in our research; then, we elucidate the intricate architecture of our network; thereafter, we showcase the datasets and results of our experiments; and finally, we provide a succinct conclusion.

The paper proceeds with a review of NIE literature, core concept definitions, CADReN's architecture, experimental datasets and results, culminating in a conclusion.

Our main contributions are:

\begin{itemize}

\item A transferable KG modeling method using CA, enabling efficient cross-graph NIE inference without retraining.

\item A novel, controllable NIE paradigm with CA as a user-network interface for flexible outcomes.

\item The introduction of \textbf{RIC200} (\textbf{R}elevant \textbf{I}nfo in \textbf{C}ontext-200) and \textbf{WK1K} (\textbf{W}i\textbf{K}ipedia-1000) datasets to foster cross-graph NIE research. (Details in section \textbf{Dataset}.)

\end{itemize}

\section{Related Works}

Node Important Estimation  began with an initial focus on structural information, further evolved to embedding-based methods capturing the rich information from KGs, and recently shifted towards more sophisticated paradigms combining these approaches with KGs and LLM. 

% \subsection{Structure-Pattern-Based Approach} PageRank (PR) \citep{Page1999ThePC} is no doubt a pioneer in NIE and it has been a successful method for estimating the importance of websites. Personalized PageRank (PPR) \citep{ppr} and Hub, Authority and Relevance Score (HAR Score) \citep{har} were then proposed to enhance PR. However, in many scenarios where KGs are being used, relying solely on node degree or other structural information can be limiting as it fails to capture the rich information. Ignoring the abundant information represented by the node descriptions leads to performance degradation in real-world applications, according to the experiments conducted in previous works \citep{geni, RGTN}.  

PageRank (PR) \citep{Page1999ThePC}, a seminal NIE technique, initially gauged the importance of web pages effectively. It was refined by Personalized PageRank (PPR) \citep{ppr} and Hub, Authority, and Relevance Score (HAR Score) \citep{har} to address its limitations. Nevertheless, these approaches, focused on node connectivity, often overlook the nuanced semantics within KGs, resulting in suboptimal performance in complex scenarios, as evidenced by empirical studies \citep{geni, RGTN}.

\subsection{Embedding-Based Approach} 
% To capture the nuance within a KG, another line of work, the embedding-based trainable frameworks arises. In the beginning, the focus is still on the structural information, such as models like node2vec \citep{node2vec}. The idea of Graph Neural Network, GNN \citep{gnn}, grew rapidly, whose principle is aggregating information from neighbors to achieve promising results. GNN-based methods gradually become an important direction of NIE, refreshing many benchmarks. With the advanced development in powerful networks like Graph Convolution Networks \citep{gcn} and Transformers \citep{gat}, embeddings are increasingly being applied in KG-related researches. For example, GENI \citep{geni} maps node features obtained by node2vec to importance values and incorporates adaptive aggregation of these values from different edge types using a graph attention network. Building upon GENI, MULTIIMPORT \citep{MULTIIMPORT} extends the exploration of learning latent node importance by incorporating external input importance signals. Both GENI and MULTIIMPORT draw inspiration from GNNs and Transformers. The advent of Transformer-based architectures, such as RGTN \citep{RGTN}, has paved the way for more effective encoding and fusion of text and structural information. However, despite their success in accurate NIE on a single KG, applying them to a new KG usually requires costly retraining. This drawback severely hinders on the application of KGs.

The advent of embedding-based frameworks marked a shift towards capturing the intricacies of KGs. Initially, methods like node2vec \citep{node2vec} still prioritized structural properties. However, the rise of Graph Neural Networks (GNN) \citep{gnn} signified a methodological leap, leveraging neighborhood aggregation to improve NIE. The continued innovation in network architectures, including Graph Convolution Networks \citep{gcn} and Transformers \citep{gat}, has seen embeddings become pivotal in KG research. For instance, GENI \citep{geni} and its successor MULTIIMPORT \citep{MULTIIMPORT} have pushed the boundaries of latent node importance identification, drawing on GNN and Transformer principles. Yet, despite their efficacy, the application of these models to new KGs often necessitates expensive retraining, limiting their practical deployment.

\subsection{Integrating KG to LLMs}

Traditional graph-based machine learning methods are facing bottlenecks in handling general knowledge and semantic understanding, necessitating the integration of LLMs with KGs. \citep{exploring}. Applications utilizing both, such as SPARQL-enhanced Question Answering \citep{sparql} and LARK's KG-based reasoning \citep{lark}, have emerged. These integrative approaches generally fall into two streams \citep{unifyingllmkg}: direct knowledge infusion during LLM training, exemplified by ERNIE \citep{ernie} and K-BERT \citep{kbert}, and prompt-based information channeling as seen in ReLMKG \citep{relmkg} and GreaseLM \citep{greaselm}. The latter, accommodating dynamic and real-time knowledge, is particularly apt for the fluid business sector. This highlights NIE's crucial role in extracting relevant information from KGs, especially given the limited context window of LLMs, to ensure that only the most critical and pertinent data is utilized for model inputs.

\section{Preliminaries}
In this section, we will provide a formal definition of the core concepts, alongside the NIE task.

\subsection{Graph}
\textbf{Definition}: A graph is a mathematical structure denoted as \(G = (V, E)\) consisting of a non-empty set \(V\) of vertices (or nodes) and a set of edges \(E\) . Vertices represent distinct entities or elements, while the edges delineate the connections or relationships between these vertices.

\subsection{Node Importance Estimation task}
\textbf{Definition}:The Node Importance Estimation task is centered on assigning an Importance Score to each node within a graph. Specifically, for a given user input \(q\) and a KG \(G\),  the goal is to identify a function \(f\) such that \(f(q,G) = I\). Here, \(I\) represents a vector wherein the i-th element signifies the Importance Score of the i-th node of \(G\). Previous work learns a function \(g\) such that \(g(G) = I\), which does not take \(q\) as an input.

\subsection{CA, BG and GT node subsets/subgraphs}
% \textbf{Definition}: CA (Contextual Anchor), BG (Background) and GT (Ground Truth) are three subsets of a graph, satisfying consecutive inclusion: CA \(\subset\) GT \(\subset\) BG. CA is the set of the nodes appeared in the user's input query \(q\). BG is the set of all the nodes in the whole graph filtering some stop words. GT is the set of all the nodes that are labeled as "important".
\textbf{Definition}: In the context of a graph, CA (\textbf{C}ontextual \textbf{A}nchor), BG (\textbf{B}ack\textbf{G}round), and GT (\textbf{G}round \textbf{T}ruth) represent three node subsets, satisfying consecutive inclusion: CA \(\subset\) GT \(\subset\) BG. The CA subset consists of nodes present in the user's input query \(q\). The GT subset comprises nodes designated as "important", which are used as training labels. The BG subset encompasses all the nodes within the graph. CA/GT/BG (sub)graphs are simply the subgraphs containing the CA/GT/BG nodes.

\section{Model Architecture}
% In this section, we first present the overall architecture of our model, and then show the details of each components. The proposed network architecture is shown in Figure~\ref{fig:pipeline}. To begin with, we extract semantic and structural features from KG separately using two different encoders. Then these features are fused for CA and BG graph respectively, allowing the fused features to incorporate both structural and semantic features. After acquiring the fused features of the CA and BG, another cross-attention mechanism is employed to facilitate the adaptive interaction between the CA features and BG features. Subsequently, a classifier is utilized to forecast the importance of each BG node. Ultimately, our model is trained employing a combination of Binary Cross-Entropy loss, semantic loss, and structural loss, as proposed by us.

In this section, we outline our model's architecture, detailed in Figure~\ref{fig:pipeline}. The process begins with separate encoders extracting semantic and structural features from the KG. These features are then fused for both CA and BG graphs, integrating structural and semantic information. A cross-attention mechanism further refines the interaction between CA and BG features. Finally, a classifier predicts the importance of each BG node, with our proposed loss function incorporating Binary Cross-Entropy loss, semantic loss, and structural loss.

\subsection{Four Branch Encoding}
% In this part, the CA and the BG graphs are encoded using a BERT Encoder and a hand-designed Structural Encoder, respectively, to obtain semantic and structural embeddings as described in the Four Branch Encoding in Figure ~\ref{fig:pipeline}.

Our model employs a dual-encoding approach, leveraging both a BERT Encoder (chosen following the setting in \citep{RGTN}) for semantic analysis and a naive Structural Encoder for structural insights. This process, termed Four Branch Encoding in Fig.~\ref{fig:pipeline}, is designed to obtain distinct semantic and structural embeddings for the CA and BG graphs.

\subsubsection{Semantic Embedding}
% To obtain the semantic embeddings of a specific node (we assume that it is the \(i\)-th node of the graph, denoted as \(node_i\)), we encode the concatenated string of all the CA entities and the entity of \(node_i\) using BERT following the previous work \citep{RGTN}. Then we extract the embedding of the first (head) and the last (tail) token of the \(node_i\)'s entity, which are concatenated as the semantic embedding of \(node_i\).

% Encoding each node along with the CA nodes is advantageous because the BERT encoding process encodes information from the CA nodes into the embedding of \(node_i\). This facilitates the model's ability to learn the relationships between nodes.

% After experimenting with various methods, including average pooling, embedding addition, and max pooling, to generate fixed-length embeddings for each node, the head-tail concatenation approach emerged as the most effective.

Semantic embedding of \(node_i\) is derived by encoding the concatenation of \(node_i\) and all CA nodes with BERT. Encoding \(node_i\) along with the CA nodes is advantageous because the BERT encoding process encodes information from the CA nodes into the embedding of \(node_i\). This facilitates learning of the relative relationships between nodes. In order to get a fix-length embedding for all the nodes, We extract and concatenate the embeddings of the first and last tokens of \(node_i\) to form its semantic representation.

\subsubsection{Structural Embedding}
% The structural embeddings are composed of several statistics to represent the structural information of a node including: the number of all its children nodes; the number of first layer children nodes; the number of steps required to reach the CA nodes, etc. 

% We selected these features because from the surveys with business analysts, these features represent the structural importance of \(node_i\) as well as the structural proximity between this node and the CA nodes. In previous works, other structural encoders are used, such as node2vec \citep{node2vec} and GNN \citep{gnn}. These methodologies facilitate the mapping of structural information onto a higher-dimensional space, thus endowing the model with enhanced representational capabilities. However, integrating relative relationships into these encoders poses notable challenges. In our devised encoder, the relative associations with CA are explicitly taken into account, thereby constituting an initial endeavor towards a CA-aware structural encoder. Subsequent experiments validate the effectiveness of this simple design in capturing relative relationships.

The structural embeddings encompass 5 key node statistics: [\#(child nodes), \#(direct child nodes), \{max,min,avg\} of steps to reach CA nodes]. These features, selected based on business analyst feedback, capture both the structural significance and proximity of \(node_i\) to CA nodes. Previous structural encoders like node2vec \citep{node2vec} and GNN \citep{gnn} facilitate the mapping of structural information onto a higher-dimensional space, thus endowing the model with enhanced representational capabilities. However, integrating relative relationships into these encoders poses notable challenges. In our devised encoder, the relative associations with CA are explicitly taken into account, thereby constituting an initial endeavor towards a CA-aware structural encoder.

\subsection{Cross-Attention Fusion}
% The second part of the model aims to integrate information: (1) between the semantic source and structural source, and (2) between the CA graph and the BG graph.

% To accomplish this, we begin by implementing cross-attention between the semantic and structural information. This involves concatenating the two embeddings and encoding them using a Transformer-like encoder. The resulting output is then split back into two separate embeddings. Likewise, we employ cross-attention between the CA and BG, employing a similar process while utilizing the semantic or structural embeddings derived from both graphs. In the end, each of the four embeddings contains information from the other three counterparts. 
% % The process of a single cross-attention is illustrated in Figure ~\ref{fig:cross_att}. 

% This module serves as an effective approach to fuse the information from four sources, enabling the model to learn the concept of ``importance" by establishing hidden relationships with the CA nodes.

% Before entering the next module, the embeddings are passed through a Reconstruction Auto-Encoder. The embeddings of some nodes are dropped randomly, then a MLP is trained to reconstruct the dropped embeddings based on the embeddings of other nodes. This mechanism helps improve CADReN's robustness.

This phase integrates semantic and structural data from both the CA and BG graphs. It employs cross-attention mechanisms, first between semantic and structural embeddings, then between the CA and BG graph embeddings. Each embedding, processed through a Transformer-like encoder, amalgamates information from the other three sources. This fusion not only enhances learning of the "importance" concept but also establishes hidden relationships with CA nodes. The embeddings undergo further refinement via a Reconstruction Auto-Encoder, which aids in model robustness by training a Multi-Layer Perceptron (MLP) to reconstruct randomly dropped node embeddings.

\subsection{Attention-based Aggregation}
% In the third part, we have implemented an Attention-based Aggregation mechanism to predict Node Importance Score (NIS) based on the embeddings obtained in the previous parts. This mechanism is illustrated in Figure ~\ref{fig:att_aggregation}. 

\begin{figure}[t]
  \centering
  \resizebox{\columnwidth}{!}{
  \includegraphics{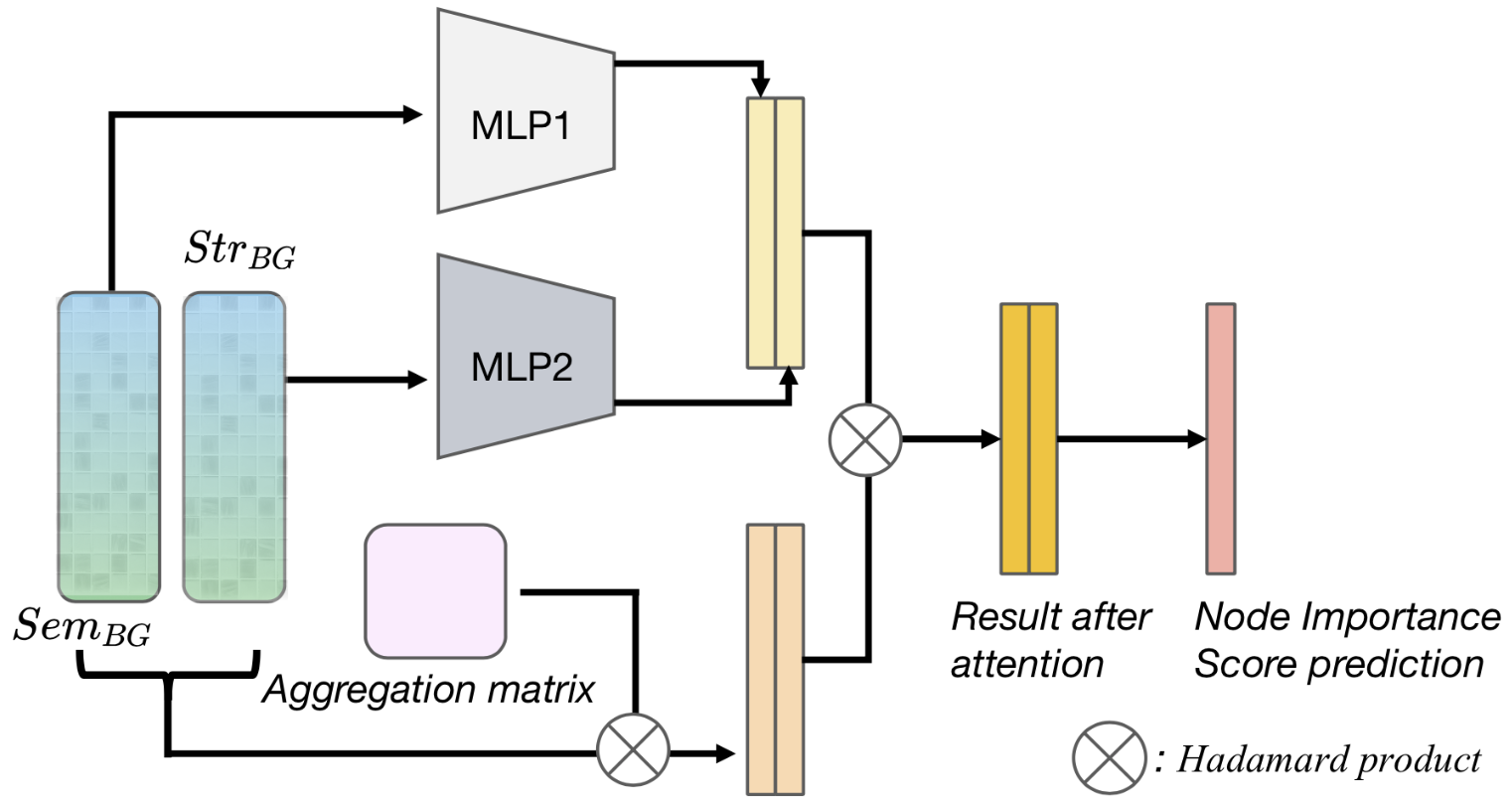}}
  \caption{Attention based Aggregation mechanism. The \textit{Aggregation matrix} contains trainable attention parameters, which are used to produce the self-attention Query that guides the prediction of Node Importance Score.}
  \label{fig:att_aggregation}
\end{figure}

The third segment of our model introduces an Attention-Based Aggregation mechanism. This component is pivotal in predicting the Node Importance Score (NIS) using the embeddings generated in the earlier stages of the model. This mechanism is illustrated in Figure~\ref{fig:att_aggregation}. 

% The key idea is to employ self-attention. Firstly, we pass the embeddings after the cross-attention module through two MLP encoders to generate the key tensor for self-attention. The embeddings are also multiplied by an ``aggregation matrix" and reshaped into a tensor with the same shape as the self-attention key tensor. This reshaped tensor serves as the query for self-attention. By performing a Hadamard product between the key tensor and the query tensor, we obtain a tensor of shape \([\#node, 2]\). Each row in this tensor contains two NIS, calculated based on the semantic and structural embeddings, respectively.

% The prediction of NIS is derived by aggregating the semantic NIS and the structural NIS, followed by the application of a softmax function.

The core principle underlying this mechanism is the utilization of self-attention. Initially, the embeddings from the cross-attention module are processed through two MLP encoders. This step generates the Key tensor for self-attention. Concurrently, the embeddings are transformed by an "aggregation matrix" and reshaped into the Query tensor that mirrors the shape of the Key tensor.

The Hadamard product between the Key and Query tensor yields a tensor of shape \([\#node, 2]\). Each row of this tensor encapsulates two NIS, one derived from semantic embeddings and the other from structural embeddings.

To finalize the prediction of NIS, the model aggregates these semantic and structural NIS values. This aggregation is then refined with a softmax function, ensuring a normalized probabilistic output for the NIS.

\subsection{Post-processing Adjustment}
In the final part, we introduce Post-processing Adjustment to further enhance the model's performance. This is achieved by calculating a weighted summation between the predicted NIS vector, the semantic similarity vector, and the structural similarity vector.

\subsubsection{Semantic Similarity Vector}
The semantic similarity vector is computed by averaging the cosine similarity between the \(node_i\)'s semantic embeddings and the embeddings of the CA nodes. The \(i\)-th element of the semantic similarity vector, denoted as \( \mathcal{S}_{sem, i}\), is calculate as follows:
\begin{equation}
\mathcal{S}_{sem, i} = \frac{\sum\limits_{j=1}^{|CA|}\langle \mathcal{E}_{sem}(node_i) | \mathcal{E}_{sem}(CA_j)\rangle)}{|CA|}
\end{equation}
where:  \(\mathcal{E}_{sem}(.) \) represents the semantic embedding obtained via BERT encoder. \(\langle.|.\rangle\) denotes the function of cosine similarity.   \(|CA|\) denotes the number of nodes in the CA set.

For nodes included in the CA graph, their semantic similarity is assigned a maximum value (1). 
% For all other nodes, their semantic similarity values range from 0 to 1. A higher value indicates a higher semantic similarity between the node and the CA nodes.

\subsubsection{Structural Similarity Vector}
The structural similarity vector is obtained using a function determined by regression. This function takes the structural features of a node as input and outputs a scalar between 0 and 1 representing the structural similarity between the node and the CA nodes. The \(node_i\)'s structural similarity \( \mathcal{S}_{str, i}\) is defined as:
\begin{equation}
\mathcal{S}_{str, i} = b + \mathcal{R}[\mathcal{E}_{str}(node_i)]^{tr}
\end{equation}

where:  \(\mathcal{E}_{str}(.)\) represents the structural embedding of a node. \(\mathcal{R}\)  and \(b\) are the regression parameters and bias respectively.

We perform the regression with 5\% randomly sampled data from the training set. The ratio between CA, GT and BG node numbers are kept during the sampling. Once the \(\mathcal{R}\) and \(b\) are determined, we fix them to calculate the structural similarity of any given node.

% It is worth to note that some transformations may be applied to the structural embeddings before performing regression. These data transformations are applied to obtain a smoother distribution of feature values, making it easier to perform regression. The transformations are manually set based on the dataset.

% As an example, when training with the RIC10K (further explanation of the dataset used is provided in the following chapter), we apply a logarithmic transformation to the ``noc" (Number Of Children) feature before regression:
% \begin{equation}
% noc' = \frac{1}{10*log_{10}(noc+1.1)}
% \end{equation}

\subsubsection{Weighted Summation}
The final NIS (\(I_{final}\)) is obtained as follows:
\begin{equation}
I_{final} = \sigma(\alpha * I_{init}+\beta * \mathcal{S}_{sem} + \gamma * \mathcal{S}_{str})
\end{equation}

where: \(\alpha\), \(\beta\) and \(\gamma\) are trainable parameters. \(\sigma(.)\) is the sigmoid function.

In this step, we refine the prediction results using the similarity vectors. The similarity vectors provide additional information about the CA nodes, enabling the model to better distinguish nodes with similar initial NIS predictions. 

\subsubsection{Loss Function}
The loss function of our model is defined as follows:
\begin{equation}
\mathcal{L}_{total} = \mathcal{B}(I_{gt}, I_{final}) + \mathcal{L}_{sem} + \mathcal{L}_{str}
\end{equation}
\begin{equation}
\mathcal{L}_{sem} = \mu * \mathcal{B}(\mathcal{S}_{sem} * I_{gt}, I_{final})
\end{equation}
\begin{equation}
\mathcal{L}_{str} = \nu * \mathcal{B}(\mathcal{S}_{str} * I_{gt}, I_{final})
\end{equation}

where: \(\mathcal{B}(.)\) is the function to calculate Binary Cross Entropy. \(I_{gt}\) and \(I_{final}\)  represent the ground truth and the prediction values of NIS. \(\mathcal{L}_{sem}\) and \(\mathcal{L}_{str}\) are loss terms weighted on semantic and structural similarities. \(\mu\) and \(\nu\) are hyperparameters.

In this loss function, we incorporate two weighted terms to prioritize the losses associated with nodes that are either semantically or structurally important. This setting strengthens the model's robustness against noise from nodes that are semantically unrelated or structurally distant from the CA nodes.

\section{Experiments}
This section describes our experiments that aim to answer the following research questions:
\begin{itemize}
    \item Cross-graph Performance:  Does CADReN outperform other approaches for cross-graph NIE tasks? Can it do zero-shot inference on different graphs without retraining?
    \item Single-graph Performance:  does our model perform on par with previous works?
    \item Impact of CA: By introducing CA, does CADReN show better flexibility and controllability in NIE tasks?
\end{itemize}

\subsection{Datasets}
Our model is designed for multi-graph scenario, for which there are no datasets readily available. We have created our own datasets, and we plan to opensource RIC200 and WK1K to the community. 

For each node inside the graphs of these datasets, it is labeled as one type among \{CA, GT, BG\}. Nodes are labeled in a way to simulate the real-world application scenario: the CA nodes given by a user reflecting his/her interest, the GT nodes showing the expected responses, and the BG nodes representing the knowledge resource. In other words, the CA and GT nodes are labeled in accordingly, we call them a "pair". It is worth mentioning that, on average, each graph has 5 pairs of (CA, GT). We use different pairs of (CA, GT) to test the model's ability to give flexible outputs.

In order to compare with previous single-graph oriented models, for most of the datasets we used, a single-graph version is constructed, by simply putting all the graphs into one giant graph. 

The datasets used are listed in Table \ref{table:datasets}:

\begin{table}[h]
\resizebox{1.00\columnwidth}{!}{
\begin{tabular}{@{}c|ccccc@{}}

\toprule
Database     & \#Edges & \#BG & \#GT & \#CA & \#Graphs \\ \midrule
FB15K-S      &  592,213& 14,591 & 1,459 & 150  & 1     \\
FB15K-M      &  3006  &  74& 7 &  5 &   197  \\ \midrule
RIC200-S     &    63,802 & 36,607 & 2,004   &  617 &   1 \\
RIC200-M     &   319 & 183 & 13 &3 & 250  \\ \midrule
RIC10K-M     &    77 & 43&10 &   3 &   10,000 \\ \midrule
WK300-S      &    97,654 & 90,746 &    1,884 & 950  &   1\\
WK300-M      &   311& 289   &    6   &  3  & 314\\ \midrule
WK1K-M       & 318 & 295 &  6  & 3 &1,024\\
 \bottomrule
\end{tabular}
}
\caption{Statistics of datasets used in our experiments. 
% \#\{BG, GT, CA\} denotes the number of nodes of \{BG, GT, CA\} sets respectively. 
All the numbers are averaged numbers. The suffix "-S/-M" represent "\textbf{S}ingle/\textbf{M}ulti-graph" version.}
\label{table:datasets}
\end{table}

\textbf{RIC10K}: a dataset containing 10k KGs covering the business landscape knowledge of different industries, which are generated based on documents like annual reports and research reports.
\textbf{RIC200}: a dataset containing 250 KGs selected from RIC10K.
\textbf{WK1K}: a dataset containing 1000 KGs that are constructed based on Wikipedia data and relevant articles, containing general knowledge across domains. 
\textbf{WK300}: a dataset containing 314 KGs selected from WK1K.
\textbf{FB15K} \citep{fb}: an open dataset containing general information across domains. Following the settings of RGTN, each node in it is accompanied with the descriptions extracted from WikiData \footnote{https://www.wikidata.org}. The NIS is represented by the node's pageview number on Wikipedia in the past 30 days. Around top-1\% (resp. top-10\%) of nodes with the highest pageview numbers are marked as the CA (resp. GT) nodes.

For the two newly proposed datasets, we give the details of their creation process here.

\textbf{RIC10K}:
Thousands of open articles are collected from the Internet. Through Named Entity Recognition and Relation Analysis, these articles are turned into 10,000 KGs, grouped by themes. In each KG, we generate some commonly asked questions (queries) with ChatGPT. The nodes mentioned in these queries are labeled as "CA" nodes. Then, a group of consulting experts labeled the nodes highly related to the given query as "GT" nodes. Overall about 7\% (resp. 23\%) of the nodes are labeled as "CA" (resp. "GT") nodes.

\textbf{WK1K}:
1,000 simulated queries are first generated with ChatGPT. For each query, its relevant articles are obtained via search engines with the query being the search input. The nodes mentioned in the queries are labeled as "CA" nodes, while the top 10\% nodes with highest word frequency in the "relevant articles" are marked as the "GT" nodes. Approximately 1\% (resp. 2\%) of the nodes are labeled as "CA" (resp. "GT") nodes.

During the experiment, when a single-graph based model (GENI, RGTN) is applied on a multi-graph dataset, the model process each graph sequentially. Multi-graph based methods (GPT-3.5, CADReN) are compatible with the single-graph setting, thus can be applied without modification.

\subsection{Baselines}
We compare our work with two previous Transformer-based methods: GENI \citep{geni}, RGTN \citep{RGTN}, as well as a representative of the generative models: GPT-3.5-Turbo \citep{gpt3} (referenced as GPT-3.5).

GENI and RGTN adopt Single-Graph Oriented Structure (SGOS), however, real-world KG datasets are composed of multiple KGs. When SGOS models are applied to these datasets, the graphs need to be aggregated into one graph first. In most scenario, this aggregation is not practical because of the size of data. Even in situations when we could aggregate the graphs, our experiments show that such work-around does not give satisfactory results (Table \ref{table:single_graph_test}). Therefore, our network is deliberately designed to adopt a Multi-Graph Oriented Structure (MGOS). To give a comprehensive comparison, our experiments cover both the single-graph and the multi-graph settings.

CA could be introduced to GPT-3.5 through prompts, while GENI and RGTN can not take CA as input by design. During the experiments of GENI and RGTN, the information from CA was carefully masked to avoid data leakage.

All the baselines were run with the same data under their default settings. The experiments are conducted on NVIDIA GeForce RTX 2080 Ti GPUs. The models are trained until convergence using the Adam Optimizer with a learning rate of 5E-3.

\subsection{Metrics}
Building upon the study conducted by GENI \citep{geni}, we employ the metrics of Normalized Discounted Cumulative Gain (\textbf{NDCG}) and Spearman’s rank correlation coefficient (\textbf{SPM}) to conduct a comprehensive evaluation of the ranking quality and importance correlation. Additionally, we introduce a novel metric called Overlap@k (\textbf{OVER}), to assess the recall of important nodes following the ranking of node importance on a dynamic range. 

\textbf{NDCG} is a commonly employed metric for evaluating the quality of rankings that takes into account the order of elements. For this specific task, we define the graded relevance values as the ground truth importance values after applying a logarithmic transformation. When presented with a list of nodes and their corresponding predicted importance scores, as well as their ground truth importance values, we sort the nodes by the predicted importance scores and take the corresponding ground truth importance at the position \(i\) as \(rel_i\). \(DCG@k\) is defined as:

\begin{equation}
DCG@k = \sum_{i=1}^{k}\frac{rel_i}{log_{2}(i+1)}
\end{equation}

The Ideal DCG (\(IDCG\)) is the DCG of the ground truth list. NormalizedDCG at position k (\(NDCG@k\)) is calculated by:

\begin{equation}
NDCG@k = \frac{DCG@k}{IDCG@k}
\end{equation}

\textbf{SPM}, or SPEARMAN, measures the correlation between the predicted NIS list \(pred\) and the ground truth list \(label\). After converting the raw values \(pred\) and \(label\) into the ranks \(R_{pred}\) and \(R_{label}\) , \(SPM\) is calculated by:

\begin{equation}
SPM = \frac{cov(R_{pred},R_{label})}{\sigma_{R_{pred}}\sigma_{R_{label}}}
\end{equation}
where: \(cov()\) is the covariance function. \(\sigma_{R_{pred}}\) and \(\sigma_{R_{label}}\) are the standard deviations of the ranks.

\textbf{OVER} is the overlap ratio of the top-m important predicted nodes (\(I_{pred}\)) and their corresponding labels (\(I_{gt}\)). Since we are evaluating a cross-graph task, the m is set dynamically to cope with graphs with different sizes. The \(OVER@k\) is attained by:

\begin{equation}
m = k* |GT|
\end{equation}
\begin{equation}
OVER@k = \frac{|I_{pred,top-m}\cap I_{gt,top-m}|}{m}
\end{equation}
where: \(|GT|\) is the number of nodes in \(GT\) set. 
% \(pred_{top-m}\) (resp. \(label_{top-m}\) ) stands for the set of the top-m important predicted nodes (resp. label nodes).

\subsection{Cross Graph Evaluation}
CADReN outperforms other approaches on multi-graph setting due to its MGOS design. The design goal of SGOS models is to learn absolute information about each node in one graph. When they are used to process multiple graphs, information from multiple graphs interfere with each other rather than complement each other. CADReN, on the other hand, with the help of CA, it can learn generalized relative relationship information from multiple graphs, leading to a significantly enhanced performance on multi-graph tasks.

Moreover, CADReN demonstrates its ability of zero-shot inference across graphs. This feature confirms that CADReN learned the transferable relative relations. Results of the experiment are organized in Table \ref{table:multi_graph_exp}.

\begin{table*}[]
\centering
\begin{tabular}{@{}c|ccc|ccc|ccc@{}}
\toprule
& \multicolumn{3}{c|}{\ul{FB15K-M}}   & \multicolumn{3}{c|}{\ul{RIC\{200$^{\dagger}$, 10K$^{\ddagger}$ \}-M}}  & \multicolumn{3}{c}{\ul{WK1K-M}} \\
Methods & NDCG                  & SPM                   & OVER & NDCG                  & SPM                   & OVER & NDCG                  & SPM                   & OVER                  \\ \midrule
\multicolumn{1}{l|}{GENI$^{\dagger}$} & \multicolumn{1}{l|}{0.7761} & \multicolumn{1}{l|}{0.4105} &  0.5168  & \multicolumn{1}{l|}{0.7825} & \multicolumn{1}{l|}{0.4277} &   0.4507   & \multicolumn{1}{l|}{0.8136} & \multicolumn{1}{l|}{0.4447} & \multicolumn{1}{l}{0.7462} \\
\multicolumn{1}{l|}{RGTN$^{\dagger}$} & \multicolumn{1}{l|}{0.8563} & \multicolumn{1}{l|}{0.4403} &   0.5502  & \multicolumn{1}{l|}{0.8228} & \multicolumn{1}{l|}{0.3247} &   0.4402   & \multicolumn{1}{l|}{0.8412} & \multicolumn{1}{l|}{0.4931} & \multicolumn{1}{l}{0.7756} \\
\multicolumn{1}{l|}{\textbf{CADReN$^{\ddagger}$}}  & \multicolumn{1}{l|}{\textbf{0.9917}} & \multicolumn{1}{l|}{\textbf{0.6294}} & \textbf{0.8988}    & \multicolumn{1}{l|}{\textbf{0.8922}} & \multicolumn{1}{l|}{\textbf{0.6232}} &  \textbf{0.8675}  & \multicolumn{1}{l|}{\textbf{0.9064}} & \multicolumn{1}{l|}{\textbf{0.6390}} & \multicolumn{1}{l}{\textbf{0.8641}} \\
\multicolumn{1}{l|}{\textbf{CADReN$^{\dagger, \triangle}$}}  & \multicolumn{1}{c|}{\ul{0.9617}} & \multicolumn{1}{l|}{\ul{0.6093}} & \ul{0.8176}   & \multicolumn{1}{c|}{\ul{0.8633}} & \multicolumn{1}{c|}{\ul{0.5899}} &  \ul{0.8412}     & \multicolumn{1}{l|}{\ul{0.9007}} & \multicolumn{1}{l|}{\ul {0.6109}} & \multicolumn{1}{l}{ \ul{0.8199}} \\
 \midrule
\end{tabular}
\caption{Evaluation results of different models across datasets under multi-graph NIE task setting. NDCG and SPM are calculated with top 20 nodes, while the k parameter of Overlap is set as 2. 
% For GENI and RGTN, only the data loading pipeline is modified to enable them to process multi-graphs, without changing their model structure. 
The results in the row of CADReN $^\triangle$ is obtained by first training CADReN on RIC10K, then inference on other datasets. Best results are in bold, second best results are underlined.}
\label{table:multi_graph_exp}
\end{table*}

\subsection{Single Graph Evaluation}
Single-graph NIE has been the center of NIE researches during a long time. In order to better compare with the previous works, CADReN is also tested under single-graph setting with baselines. Experiment results are organized in Table \ref{table:single_graph_test}. The results show that, even though CADReN is not built upon single-graph scenario, it still matches the performance of previous works, getting the best or second best outcomes in most tests.

\begin{table*}[]
\centering
\begin{tabular}{@{}c|ccc|ccc|ccc @{}}
\toprule
    & \multicolumn{3}{c|}{\ul{FB15K-S}}  & \multicolumn{3}{c|}{\ul{RIC200-S}}  & \multicolumn{3}{c}{\ul{WK300-S}}  \\
Methods & NDCG                  & SPM                   & OVER & NDCG                  & SPM                   & OVER & NDCG                  & SPM                   & OVER                  \\ \midrule
\multicolumn{1}{l|}{GENI} & \multicolumn{1}{l|}{0.9191} & \multicolumn{1}{l|}{0.7520} &   0.3901   & \multicolumn{1}{l|}{\textbf{0.7095}} & \multicolumn{1}{l|}{0.4231} &  0.2412   & \multicolumn{1}{l|}{\textbf{0.5899}} & \multicolumn{1}{l|}{0.2326} & \textbf{0.1700} \\
\multicolumn{1}{l|}{RGTN}  & \multicolumn{1}{l|}{\textbf{0.9550}} & \multicolumn{1}{l|}{\textbf{0.8007}} &  \textbf{0.4720}   & \multicolumn{1}{l|}{\ul{0.6622}} & \multicolumn{1}{l|}{\ul{0.4387}} & \ul{0.2500}  & \multicolumn{1}{l|}{0.5257} & \multicolumn{1}{l|}{\textbf{0.2741}} & 0.1600 \\
\multicolumn{1}{l|}{\textbf{CADReN}}  & \multicolumn{1}{l|}{\ul{0.9322}} & \multicolumn{1}{l|}{\ul{0.7743}} &  \ul{0.4172}   & \multicolumn{1}{l|}{0.6321} & \multicolumn{1}{l|}{\textbf{0.4778}} & \textbf{0.2612}  & \multicolumn{1}{l|}{\ul{0.5311}} & \multicolumn{1}{l|}{\ul{0.2601}} & \ul{0.1612} \\\midrule
\end{tabular}
\caption{Evaluation results of different models on single-graph datasets. 
% All the models are trained on each datasets mentioned from the very beginning until convergence. 
NDCG and SPM are calculated on the top 100 nodes, while the k parameter of Overlap is set as 2. 
% From FB15K to WK300, the dataset become more and more heterogeneous so the performance of all the methods drops, showing that aggregating the small graphs is not an ideal way to do multi-graph NIE task. 
CADReN achieves similar performance on single-graph NIE compared with previous works even though it is not specifically designed for it. Best results are in bold, and second best results are underlined.
}
\label{table:single_graph_test}
\end{table*}

\subsection{Effectiveness of CA}
The introduction of the CA allows users to interact with the NIE network, leading to more accurate and more flexible NIE predictions. To demonstrate this feature, we apply NIE with fixed BG nodes while altering the (CA, GT) pairs. CADReN successfully captures this change and gives prediction accordingly, while previous works can not adapt to the change of context. One qualitative result is shown in Fig.~\ref{fig:ex_1}. More results in \textbf{Appendix A}.

\begin{figure}[!h]
  \centering
  \resizebox{\columnwidth}{!}{
  \includegraphics{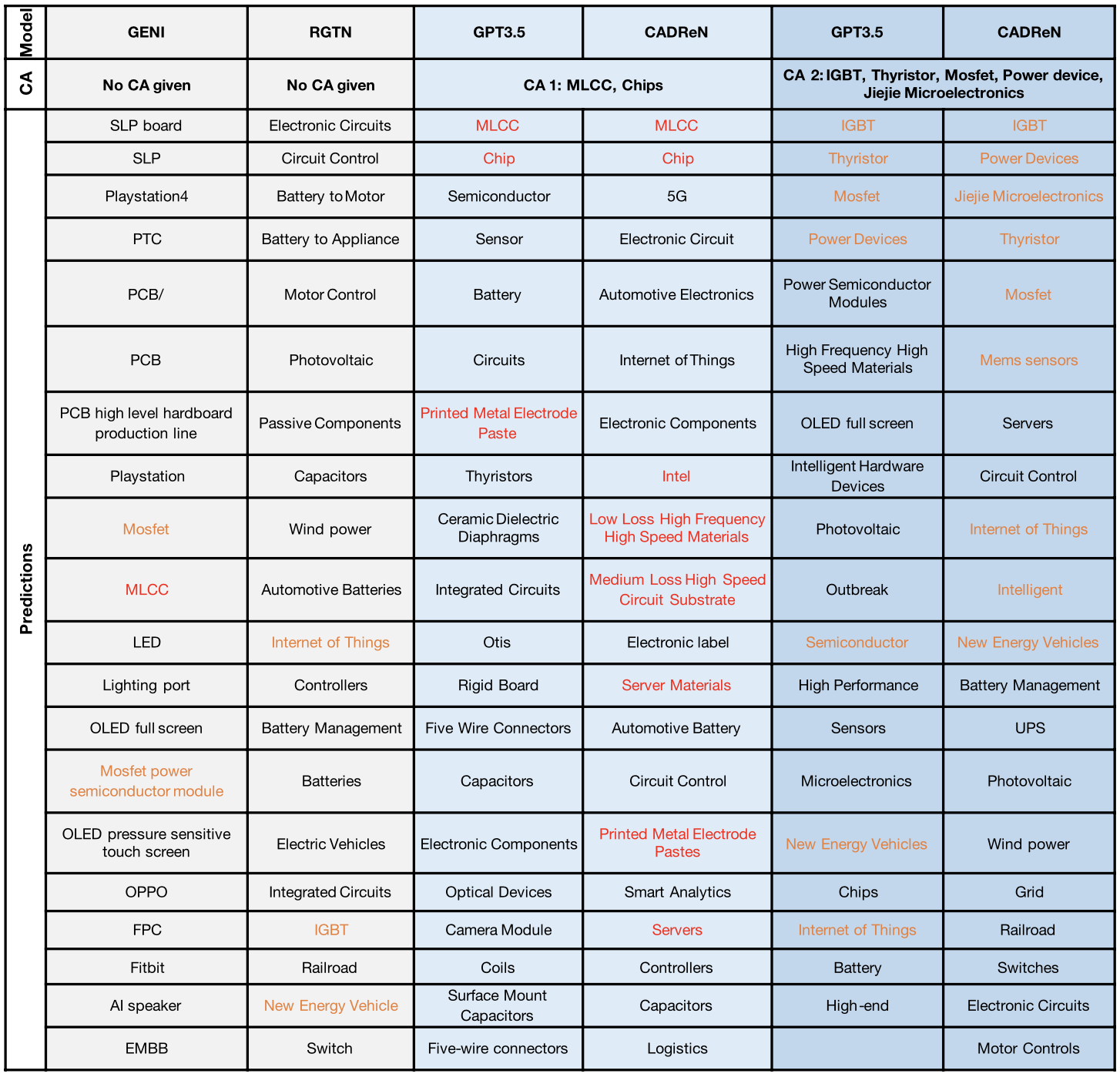}}
  \caption{Top 20 nodes with highest NIS predicted. Red (resp. orange) nodes are GT nodes corresponding to CA 1 (resp. CA 2) nodes. }
  \label{fig:ex_1}
\end{figure}

\subsection{Effectiveness of Structural Information}
LLMs are powerful for textual information analysis, it is natural to use LLM for NIE tasks directly. However, due to the lack of structural information and of up-to-date information,
% especially the structural information showing how relationships among entities evolve when real-world events occur (e.g. disruptive events, new invention etc.),  
GPT-3.5 shows less ideal performance, as shown in Table \ref{table:structure_exp}.

\begin{table}[h!]
\centering
\small
\resizebox{1.00\columnwidth}{!}{
\begin{tabular}{c|ccc|ccc}
\toprule
    & \multicolumn{3}{c|}{\ul{RIC200-M}}  & \multicolumn{3}{c}{\ul{WK300-M}}   \\
Methods & \scriptsize{NDCG} & \scriptsize{SPM} & \scriptsize{OVER} & \scriptsize{NDCG} & \scriptsize{SPM} & \scriptsize{OVER}    \\ \midrule
GPT-3.5 &   \multicolumn{1}{c|}{0.41}  &   \multicolumn{1}{c|}{0.51}   &    0.21   &   \multicolumn{1}{c|}{0.61}   &   \multicolumn{1}{l|}{0.55}  &    0.45   \\
\textbf{CADReN}  & \multicolumn{1}{c|}{0.87}  & \multicolumn{1}{c|}{0.61} &  0.85 &   \multicolumn{1}{c|}{0.92} &   \multicolumn{1}{l|}{0.63}   &   0.87    \\ \midrule 
\end{tabular}
}
\caption{GPT-3.5's ability on NIE task is not satisfactory due to the lack of structural information and of up-to-date information.}
\label{table:structure_exp}
\end{table}

\subsection{Ablation Tests}
Additional ablation tests are carried out to evaluate the effectiveness of the mechanisms that we proposed: the \textbf{C}ontextual \textbf{A}nchor (CA), the \textbf{A}ttention-bassed \textbf{A}ggregation (AA), the \textbf{A}uto-\textbf{E}ncoder (AE) and the \textbf{P}ost-\textbf{P}rocessing mechanism (PP). We measure the performance of CADReN on RIC10K with these modules partially disabled. Experiments confirm the effectiveness of these components. Results are organized in Table \ref{table:ablation}.

\begin{table}[]
\centering
\small
\begin{tabular}{c|ccc}
\toprule
            & NDCG & SPM & OVER \\ \midrule
w/o CA & \multicolumn{1}{l|}{0.6968} & \multicolumn{1}{l|}{0.3211} & 0.1275  \\
w/o AA & \multicolumn{1}{l|}{0.7338} & \multicolumn{1}{l|}{0.5363} & 0.8095  \\
w/o AE & \multicolumn{1}{l|}{0.8647} & \multicolumn{1}{l|}{0.6071} & 0.7979    \\
w/o PP & \multicolumn{1}{l|}{0.8823} & \multicolumn{1}{l|}{0.6121} &  0.8207   \\
\textbf{CADReN}   & \multicolumn{1}{l|}{0.9064} & \multicolumn{1}{l|}{0.6390} & \multicolumn{1}{l}{0.8641}  \\ \midrule
\end{tabular}
\caption{Ablation test: each proposed component of CADReN helps to improve the overall performance.
% , with the introduction of CA bringing the most significant improvement. 
% * The convergence of the model is not guaranteed without PP, the table shows the converged results.
}
\label{table:ablation}
\end{table}

\section{Conclusion}

In conclusion, our method is the first work to emphasize the relative relationship between a Contextual Anchor and other nodes within a Knowledge Graph using a Transformer-based architecture, while utilizing both structural and semantic information, to tackle the cross-graph Node Importance Estimation task. Our approach outperforms existing methods on cross-graph NIE setting and achieves similar performances on single-graph NIE setting. The introduction of CA enables the model to give flexible and accurate predictions.

To further enhance performance, future research could delve into the exploration of novel encoding mechanisms to generate superior embeddings. Specifically, in the case of structural embeddings, there is ample room for improvement. Neural networks, such as Graph Neural Networks, hold promise in providing more detailed structural information. However, a challenge persists in accurately representing the relative distance between the Contextual Anchor and the nodes in background graph. Addressing this issue is of utmost importance for forthcoming researches in this field.

% Entries for the entire Anthology, followed by custom entries
% \bibliography{anthology, NAACL_CADReN}
\bibliography{NAACL_CADReN}
% \bibliography{anthology,custom}

\section{Appendix}

\appendix

\section{More results explained in details}
\subsection{Results of \textit{Effectiveness of CA} experiment}
Here we show the results of different models applied on same BG graphs while altering the CA and GT nodes in figure \ref{fig:appen_1} and figure \ref{fig:appen_2}. The nodes marked in red (resp. orange) are the nodes contained in the \(GT_1\) (resp. \(GT_2\)) set related to the \(CA_1\) (resp. \(CA_2\)) nodes.

\subsubsection{Comparison between the gray and blue columns} 
\begin{figure*}[!h]
  \centering
  \resizebox{\textwidth}{!}{
  \includegraphics{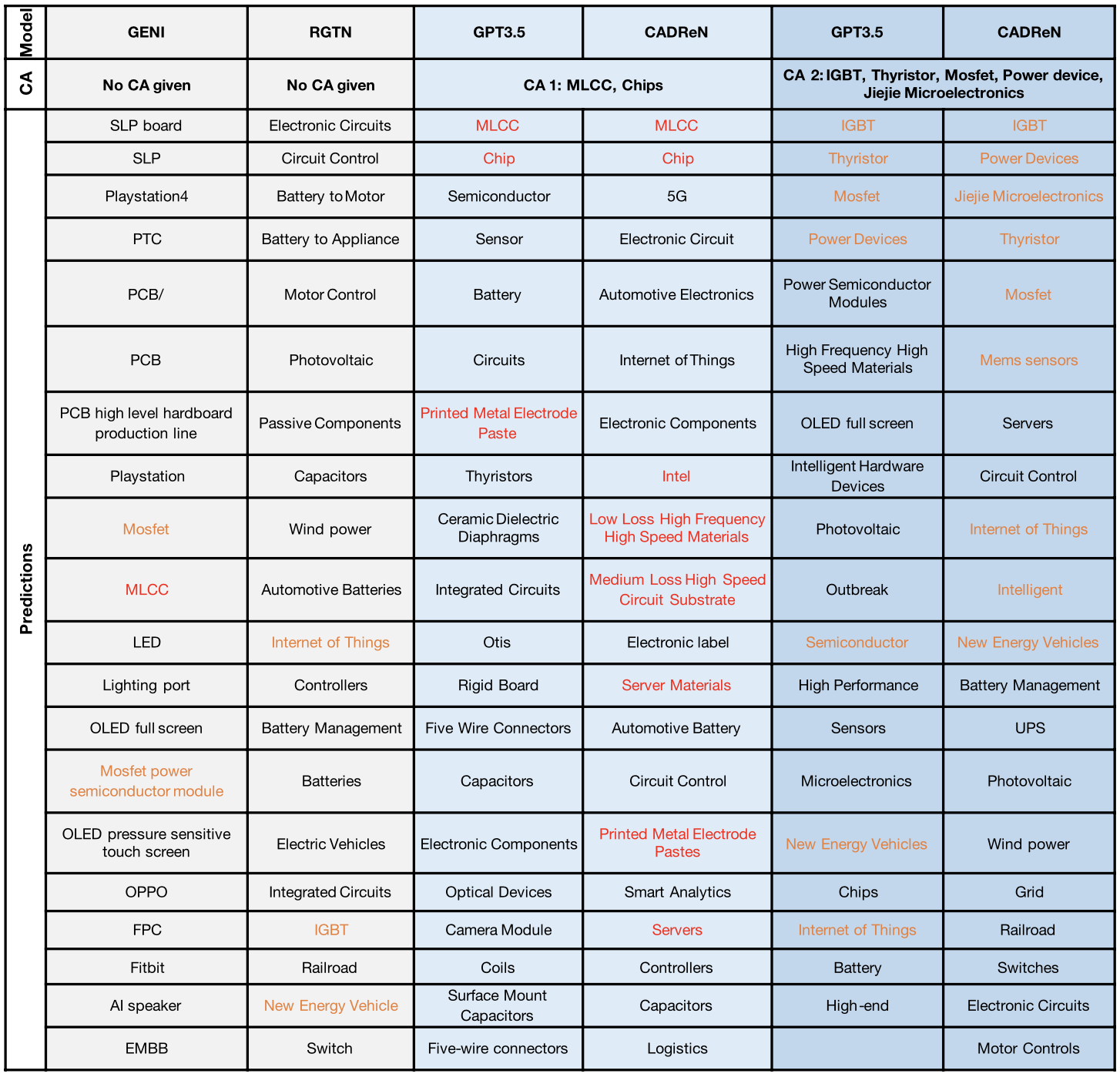}}
  \caption{Results of experiment on BG No. 1610703}
  \label{fig:appen_2}
\end{figure*}

GENI and RGTN could not take CAs as input, therefore, their predictions are static and not flexible, usually including the generally ``popular" nodes (e.g. PlayStation 4) or the acronyms linked to lots of nodes (e.g. DMC and 6F) but are not necessarily related to the user's interest. On the other hand, GPT-3.5 and CADReN could generate predictions
\begin{figure*}[!h]
  \centering
  \resizebox{\textwidth}{!}{
  \includegraphics{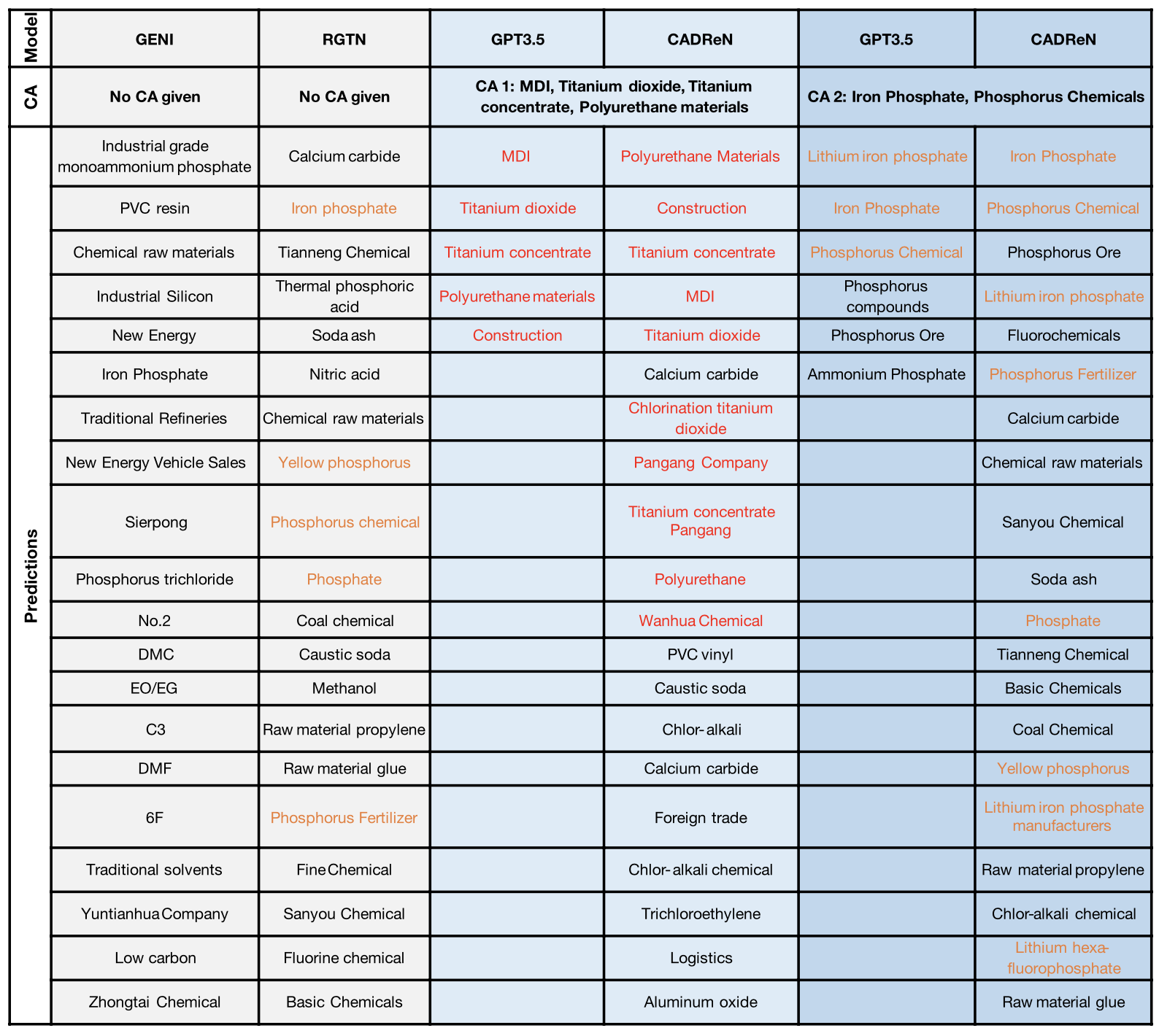}}
  \caption{Results of experiment on BG No. 1608708}
  \label{fig:appen_1}
\end{figure*}
according to  different CAs. In figure \ref{fig:appen_1}, CADReN successfully gives the predictions related to \textit{Titanium} and \textit{Phosphorus chemicals} respectively, and in the example of figure \ref{fig:appen_2}, CADReN could distinguish whether the user focuses on \textit{Chips} or \textit{Thyristors}.

\subsubsection{Comparison between light blue and dark blue columns} CADReN's predictions are more stable reasonable than the ones given by GPT-3.5. As shown in the figure \ref{fig:appen_1}, GPT-3.5 failed to provide a comprehensive prediction likely due to the lack of the niche knowledge of \textit{MDI} or \textit{Titanium dioxide}. As comparison, CADReN gives better prediction covering almost all the \(GT\) nodes among top-20 predictions because it can effectively leverage the structural information in KG as 
from semantic perspective, GPT-3.5 is superior than BERT.

\section{Prompts used during the experiments of GPT-3.5}

\mbox{}
\par\noindent\rule{\linewidth}{0.4pt}

``\textbf{role}":``system",``\textbf{content}":``you are an amazing analyst". ``\textbf{role}":``user",``\textbf{content}":`` Please select top 20 important words based on the key words  from a given set of background words. For the important words, please also provide a score (0 to 1). Output should be like word \textbackslash t score. Thank you. 

Key words:

```
\textit{\(CA_1\ \) AND \(\ CA_2\ \) AND \(\ CA_3\)} 
'''

A set of background words: 

```
\textit{\(BG_1,\ BG_2,\ BG_3,\ BG_4,\ BG_5,\ BG_6,\ ...\)} 
'''
\par\noindent\rule{\linewidth}{0.4pt}
\mbox{}

The \(CA_i\) and \(BG_j\) are filled with actual node entities during the experiments.

\end{document}